\documentclass[10pt,twocolumn,letterpaper]{article}

\usepackage[pagenumbers]{cvpr} 

\definecolor{cvprblue}{rgb}{0.21,0.49,0.74}
\usepackage{graphicx}
\usepackage{amssymb}
\usepackage{xcolor}
\usepackage[misc]{ifsym}
\usepackage{array}
\usepackage[pagebackref,breaklinks,colorlinks,allcolors=cvprblue,urlcolor=magenta]{hyperref}
\newcommand\blfootnote[1]{%
    \begingroup
    \renewcommand\thefootnote{}\footnote{#1}%
    \addtocounter{footnote}{-1}%
    \endgroup
}
\usepackage[hang]{footmisc}

\title{EchoMimicV3: \textcolor{cyan}{1.3B} Parameters are All You Need for \textcolor{cyan}{Unified} \textcolor{CarnationPink}{Multi-Modal} and \textcolor{CarnationPink}{Multi-Task} Human Animation}
\author{
    Rang Meng\textsuperscript{\rm \dag}\textsuperscript{\rm \ddag} \quad
    Yan Wang \quad 
    Weipeng Wu \quad  
    Ruobing Zheng \quad  
    Yuming Li\textsuperscript{\rm \ddag} \quad 
    Chenguang Ma\textsuperscript{\rm \ddag}
\vspace{2mm}
\\
\fontsize{11pt}{13pt}\selectfont{Terminal Technology Department, Alipay, Ant Group}
\\
{\tt\small
    \fontsize{11pt}{13pt}\selectfont{\{mengrang.mr, luoque.lym, chenguang.mcg\}@antgroup.com}}
    \vspace{1.5mm}
    \\
    \fontsize{12pt}{13pt}\selectfont{code: \href{https://github.com/antgroup/echomimic_v3}{\textcolor{cyan}{https://github.com/antgroup/echomimic\_v3}}}
}

\begin{document}
\maketitle
{
\blfootnote{
    {\rm \dag} Core contributor\\
    {\rm \ddag} Corresponding author}
}

\begin{abstract}
Recent work on human animation usually incorporates large-scale video models, thereby achieving more vivid performance. However, the practical use of such methods is hindered by the slow inference speed and high computational demands.
Moreover, traditional work typically employs 
separate models for each animation task, increasing costs in multi-task scenarios and worsening the dilemma. To address these limitations, we introduce EchoMimicV3, an efficient framework that unifies multi-task and multi-modal human animation. At the core of \textbf{EchoMimicV3} lies a threefold design: a \textbf{Soup-of-Tasks} paradigm, a \textbf{Soup-of-Modals} paradigm, and a novel \textbf{training and inference strategy}. The Soup-of-Tasks leverages multi-task mask inputs and a counter-intuitive task allocation strategy to achieve multi-task gains without multi-model pains. Meanwhile, the Soup-of-Modals introduces a Coupled-Decoupled Multi-Modal Cross Attention module to inject multi-modal conditions, complemented by a Multi-Modal Timestep Phase-aware  Dynamical Allocation mechanism to modulate multi-modal mixtures. Besides, we propose Negative Direct Preference Optimization, Phase-aware Negative Classifier-Free Guidance (CFG), and Long Video CFG, which ensure stable training and inference. Extensive experiments and analyses demonstrate that EchoMimicV3, with a minimal model size of 1.3 billion parameters, achieves competitive performance in both quantitative and qualitative evaluations.
\end{abstract}    
\section{Introduction}
\begin{quote}
\noindent\textbf{``\textit{Faster, Higher, Stronger -- Together}''}

    \hfill --- \textit{The Olympic Motto}
\end{quote}

Recent advancements in human animation have been boosted by large-scale video diffusion models (LVDM)~\cite{wang2025fantasytalking,chen2025hunyuanvideoavatar,yang2024cogvideox,wan2025wan,kong2024hunyuanvideo,wei2025mocha,cui2024hallo3,lin2025omnihuman}. While the adoption of LVDMs has led to higher quality and stronger generalization in human animation, it also introduces notable challenges: 1) prohibitive training costs and slow inference speeds due to LVDM's massive parameter scale; 2) complex model routing caused by the need for separate expert LVDMs for different animation tasks (e.g., text-to-video, image-to-video, lip-sync). Unfortunately, the latter challenge further aggravates the inefficiencies of the former.

This raises a critical question: \textit{\textbf{How can we achieve Faster inference, Higher quality, Stronger generalization, and unified support for multi-task human animation Together within a single model?}}

Analyzing from first principles, we identify that the parameter inflation in existing LVDMs is the root cause of this question we aim to answer in this paper. Hence, we employ a compact video diffusion model (CVDM) as our straightforward backbone. However, CVDMs inherently compromise on quality, generalization, multi-task unification, and multi-modal processing compared to their larger counterparts. To overcome these inherent issues of CVDM, we introduce our novel framework, EchoMimicV3, which incorporates three key innovations:

\noindent\textbf{Soup-of-Tasks for Multi-Task Unification.} We reformulate diverse animation tasks from a novel spatiotemporal reconstruction perspective akin to Masked Autoencoders (MAE)~\cite{kong2025let}. Specifically, 
lip-syncing is recast as mouth spatial region reconstruction, while Text-to-Video (T2V), Image-to-Video (I2V), and First-Last-Frame-to-Video (FLF2V) can be viewed as intermediate temporal frames reconstruction. EchoMimicV3 unifies these tasks by fully exploiting the common masked sequence input inherent in mainstream video diffusion models.
By doing so, step-wise diffusion model and patch-wise reconstruction elegantly converge without painful trade-offs. 

To dynamically allocate various tasks in the training, a counter-intuitive "hard-to-easy" training strategy is employed: we first train on complex tasks (I2V/FLF2V) to fully leverage pretrained knowledge, then incorporate simpler tasks (e.g., lip-syncing) using Exponential Moving Average (EMA) to implicitly mix task expert models. This approach enables seamless cross-task knowledge transfer and prevents catastrophic forgetting within a single model. We term this multi-task learning as Soup-of-Tasks.

\noindent\textbf{Soup-of-Modals for Mixture of Multiple Modal-Experts.}
We introduce a novel Soup-of-Modals paradigm to enhance multi-modal processing in lightweight models, following a \textbf{\textit{couple, decouple, mix}} workflow: 1) Couple: A shared query MLP couples all modalities;
2) Decouple: Modal-specific cross-attention modules inject modality-specific keys and values;
3) Mix: Modal experts are dynamically fused via \textbf{Multi-Modal timestep Phase-aware Dynamic Allocation (Multi-Modal PhDA)}.

Inspired by the PhD Loss in EchoMimicV2~\cite{meng2025echomimicv2}, our motivation of the Multi-Modal PhDA stems from the observation that \textbf{\textit{different modals exhibit varying levels of importance across different timestep phases}}. Specifically, text conditions maintain consistent importance throughout phases, image conditions are most influential during the early and middle timestep phases, and audio conditions are particularly relevant in the initial phase. The Multi-Modal PhD allocates weights to each modal expert branch based on this Phase-specific Modal Importance Law, fusing them through linear combination.

\noindent\textbf{Novel Training and Inference Strategy.} We propose a novel training strategy to effectively integrate the aforementioned Soup-of-Tasks and Soup-of-Modals paradigms. Traditional post-training methods like Direct Preference Optimization (DPO)~\cite{wallace2024diffusion} reject undesired distributions with preference data but suffer from high computational costs, limited generalization, and sensitivity to preference data quality. We propose \textbf{Negative DPO}, which uses pairing-free negative samples to reject the distribution of preference negative data. By interleaving Negative DPO with Supervised Fine-Tuning (SFT), we dynamically mitigate spatial inconsistencies (e.g., identity preservation issues) and temporal artifacts (e.g., color shifts) in Negative DPO-SFT cycle training. Notably, our Negative DPO simplifies the conventional DPO pipeline while achieving effective performance.

Through this training strategy, we observed that the model effectively activates the negative sample rejection mechanism during inference. Consequently, we introduced a novel inference approach termed \textbf{timestep Phase-aware Negative classifier-free Guidance (PNG)}, which applies weighted negative prompts at specific diffusion timesteps to suppress undesired artifacts, such as unnatural gestures and color inconsistencies. Additionally, we incorporated an enhanced long-video generation technique, significantly improving the quality of long-video outputs.

Extensive experiments demonstrate that EchoMimicV3 achieves competitive performance compared to state-of-the-art methods. Moreover, EchoMimicV3 is compatible with diverse scenarios, including podcasts, karaoke, and dynamic scenes, while maintaining computational efficiency. We will release our code for community use. 
In summary, our contributions are as follows:

\begin{figure*}[t!]
\begin{center}
\includegraphics[width=1\linewidth]{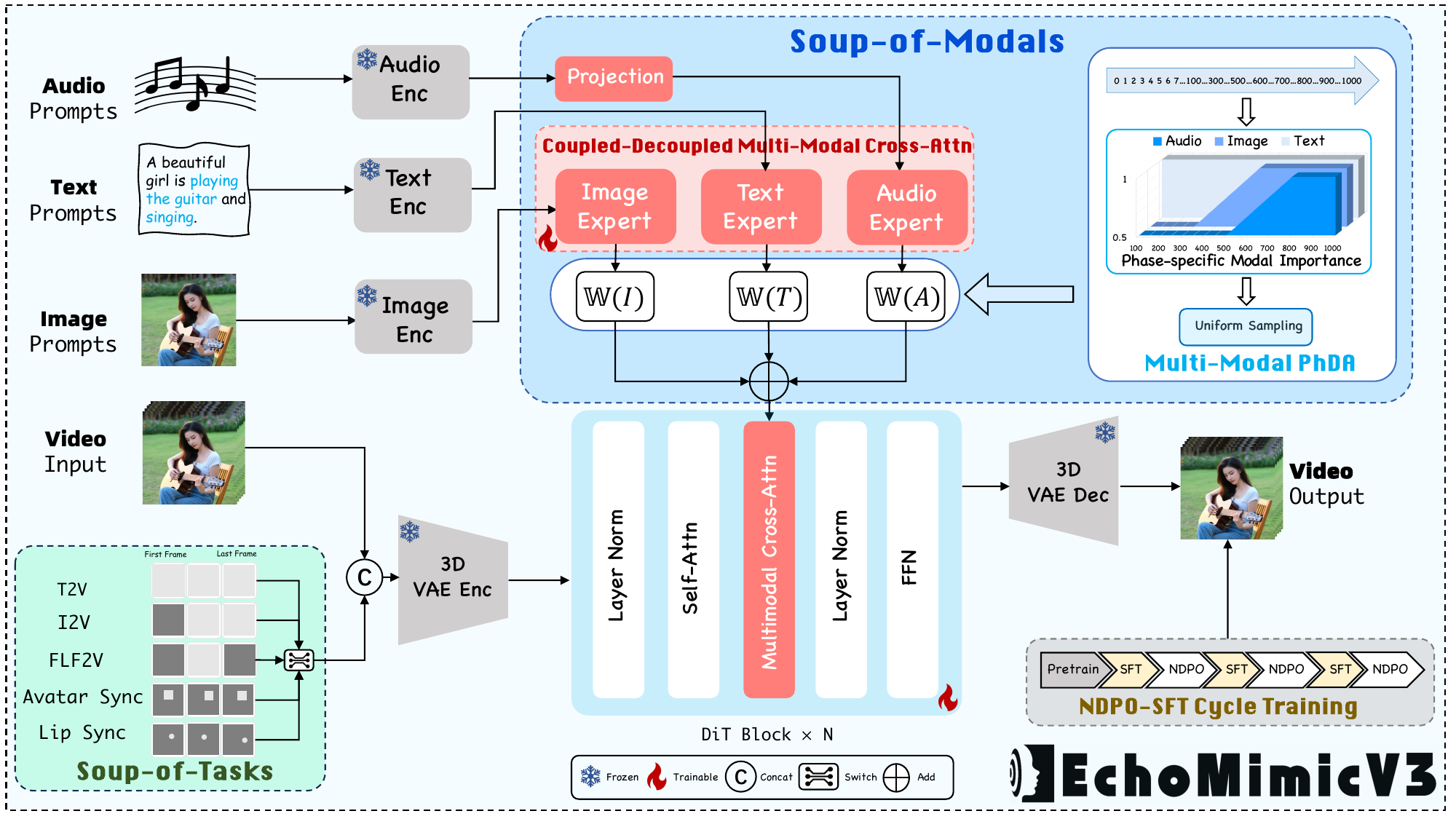}
\end{center}
\vspace{-1em}
   \caption{The overall training pipeline of EchoMimicV3.}
\vspace{-1em}
\label{fig:pipeline}
\end{figure*}

\begin{itemize}
\item We propose a lightweight framework for human animation that achieves multi-task versatility and multi-modal learning, enabling vivid performance.

\item We introduce the Soup-of-Tasks paradigm, including unified spatial-temporal masked reconstruction inputs and a counter-intuitive inter-tasks training schedule.

\item We present the Soup-of-Modals paradigm, featuring a Coupled-Decoupled Multi-Modal Cross Attention module and a Multi-Modal PhDA, for interaction and fusion across multiple modals.

\item We propose a novel Negative DPO-SFT cycle training strategy that embeds pairing-free DPO into SFT training, enabling dynamic rejection of undesirable distributions.

\item We introduce a Phase-aware Negative-enhanced CFG and Long Video CFG for vivid and long-term video inference, respectively.

\item Our EchoMimicV3, with 1.3B parameters, achieves strong competitiveness against SOTA models with ten times the parameter count, as demonstrated by both quantitative and qualitative evaluations.
\end{itemize}
\section{EchoMimicV3}
\subsection{Overview}
\noindent\textbf{Pipeline}. 
In this section, we present EchoMimicV3, a unified framework for multi-task and multi-modal human animation, as shown in Fig. \ref{fig:pipeline}.
EchoMimicV3 generates talking human videos with conditions of the reference image, audio, and text prompt, without cumbersome conditions such as predefined 2D or 3D poses. To unify diverse tasks, we utilize a new Soup-of-Task paradigm (Sec 3.2). Additionally, to enhance the multi-modal capabilities of lightweight models with 1.3B parameters, we propose a novel Soup-of-Modal paradigm (Sec 3.3). Furthermore, to ensure the stability of our framework containing various tasks and multiple modals, we propose a training strategy, in which a new Negative DPO is injected into the SFT process for dynamically rejecting undesirable results. Correspondingly, we introduce a Phase-aware Negative CFG (PNG) to enhance the rejection of negative results for inference (Sec 3.4).

\subsection{Soup-of-Tasks}
\vspace{-0.02in}
\noindent\textbf{Spatial-temporal masked reconstruction.} In the transformer of Wan-Fun series video model~\cite{wallace2024diffusion}, a corresponding 0-1 masked sequence
is concatenated with the video latent as the input.
Inspired by MAE~\cite{he2022masked}, we view diverse tasks from the spatial-temporal reconstruction perspective: diverse tasks can be viewed as employing distinct masking strategies, with variations existing solely in the mask inputs. Specifically, the mask input for the audio-driven Text-to-video (T2V), Image-to-Video (I2V), First-Last-Frame-to-Video (FLF2V), and lip syncing are designed as $\mathsf{M}_{T2V}$, $\mathsf{M}_{I2V}$, $\mathsf{M}_{FLF2V}$ and $\mathsf{M}_{lip}$, respectively, as illustrated in Fig. \ref{fig:pipeline}, where each task is encoded with a unique 0-1 sequence pattern. This design facilitates the integration of multiple tasks into a unified model without necessitating architectural modification.

\noindent\textbf{Soup-of-Tasks training strategy} includes a counter-intuitive task schedule and implicit task mixture approach.
Intuitively, curriculum learning involves progressing from easier (task with the highest mask ratio) to more challenging tasks (task with the lower mask ratio). However, in contrast, we adopt a counter-intuitive task schedule: we begin with the most difficult tasks with the highest masking ratio and gradually incorporate other simpler tasks. The reason for doing so is mainly that the challenging tasks align more closely with the task paradigm of the pretrained model.

Furthermore, we adopt an inter-task EMA (Exponential Moving Average) training strategy to implicitly mix various tasks into Soup-of-Tasks: we first train the anchor task the same as pretraining (with a high masking ratio), and then incrementally integrate other tasks with decreasing masking ratios through EMA. By doing so, we ensure full exploitation of pretrained knowledge while avoiding catastrophic forgetting, enabling the model to adapt to new tasks without losing prior task performance.

\subsection{Soup-of-Modals}
\vspace{-0.05in}
Soup-of-Modals is designed to amortize the injection, fusion and training of multi-modal conditions on multiple timestep phases, which contains a Coupled-Decoupled Multi-Modal Cross Attention (CDCA) module, and a timestep Phase-aware Dynamical Allocation (Multi-Modal PhDA) mechanism. 

\noindent\textbf{CDCA module}.
The text, audio, and image prompts are encoded by umT5, an audio extractor, and CLIP, respectively, into text, audio, and image features ($c_t$, $c_a$, $c_i$) respectively.
Given $z$ and $z_{o}$ as the input and output of Coupled-Decoupled Multi-Modal Cross Attention (CDCA). The audio, text, and image features are injected into CDCA as follows:
\begin{equation}
\begin{aligned}
&z_{o}= \sum_{c \in \{t, i, a\}}\mathbb{W}(c, \tau)\cdot\mathbb{CA}_c(Q_{shared}, K(c), V(c))
\end{aligned}
\end{equation} 
where $\mathbb{W}(c, \tau)$ denotes the $\tau$ timestep-related weights for the $c$ modal experts (cross attention) $\mathbb{CA}$. 
The CDCA module is designed similarly to IP-Adapter for multi-modal injection and fusion. Specifically, we project the conditions separately as keys and values, i.e., $K(c), V(c)$, while sharing the same query $Q_{shared}$ for performing text, audio, and image experts (cross attention). The outputs from these operations are weighted by $\mathbb{W}(c, \tau)$ and then summed.

\noindent\textbf{Multi-Modal PhDA} is designed for allocating the $\mathbb{W}(c, \tau)$ for different modals, according to the timestep $\tau$ in each iteration. The weights are sampled from a meta distribution, based on the Phase-specific Modal Importance inspired by the PhD Loss~\cite{meng2025echomimicv2}. The weights for different modals in different timestep phases can be calculated as follows:
\begin{equation}
\begin{aligned}
\mathbb{W}(c, \tau) = 
\begin{cases} 
0.5 & \text{if } \tau < \mathcal{B}^1_c, \\
m\cdot\tau+b & \text{if } \mathcal{B}^1_c \leq x < \mathcal{B}^2_c, \\
1 & \text{if } x \geq \mathcal{B}^2_c.
\end{cases}
\end{aligned}
\end{equation} 
where $\{\mathcal{B}^1_c, \mathcal{B}^2_c\} \in [0,1000]$ denote the left and right critical timestep at which modal $c$ begins to increase with timestep, and 
$m$ and $b$ represent the slope and intercept of the transition phase, respectively.

\noindent\textbf{Audio injection.} Given the audio input $ c_a $, we adopt an audio encoder $ E_a$ to extract audio features and an MLP as an audio projection module. Due to the temporal downsampling ratio $r$ introduced by the VAE in DiT, a single latent frame corresponds to $r$ audio feature tokens along the temporal dimension.

\noindent\textbf{Audio Segments.}
To this end, we adopt an audio segmentation strategy to achieve temporal alignment between audio features and latent frames, followed by implementing latent frame-wise cross-attention. 
Denote the audio embeddings as $\{c_a^1, c_a^2, \dots, c_a^{t_a}\}$, where $ t_a $ represents the length of the audio sequence. The audio embeddings are evenly divided into segments, each containing $ r \times \alpha $ features, where $\alpha$ denotes the number of audio embedding features corresponding to the duration of one video frame.

\noindent\textbf{Segments-wise audio-frame alignment.} Next, we identify the center feature of each segment over its duration, denoted as $\{c_a^{m_1}, c_a^{m_2}, \dots, c_a^{m_\tau}\}$, where $\tau$ is the temporal length of the latent. Subsequently, we extend both forward and backward by $ r + e $ features from each center feature, resulting in latent frame-wise audio embedding segments $\{s_a^1, s_a^2, \dots, s_a^\tau\}$, where $ e $ represents the overlap added for smoothness. Finally, the audio segments are injected into DiT via the audio modal expert (Audio Cross Attention). Note that the audio expert's output is modulated by a binary facial region hard attention mask $\mathbb{M}_{face} \in\{0,1\}$ to enhance the naturalness of lip synchronization and facial expressions.

\subsection{Training Strategy}
\noindent\textbf{Negative Direct Preference Optimization (Negative DPO)} leverages suboptimal negative samples generated from intermediate checkpoints of the Supervised Fine-Tuning (SFT) to iteratively refine the model. Specifically, 
during SFT, denote intermediate checkpoints as $ \{ \mathcal{M}_\tau \mid \tau \in I \}$, where $\tau$ is the number of iterations. We sample checkpoints $ \{ \mathcal{M}_{s_i} \mid {s_i} \in I \}$, where $s_i$ denotes the number of iterations for the $i$ training stage. Note that each training stage has the same issue. For the $i$ training stage, we employ $\mathcal{M}_{s_i}$ as our reference model to generate videos $\mathcal{D}_{s_i}$, then we identify negative samples from $\mathcal{D}_{s_i}$, and then annotate the issue based on user feedback, omitting the pairing of $\langle p,  y^-, y^+ \rangle$. The negative preference data for stage $i$ can be denoted as $\langle p^-, y^- \rangle\in \mathcal{D}_{s_i}$.

Unlike traditional DPO, our optimization objective is designed to only minimize the generation probability of these negative samples, to penalize the model's tendency to undesirable distribution. Our optimization objective for stage $i$ is:
\begin{equation}
\begin{aligned}
\mathcal{L}^{i}_{NDPO}(\theta) &= 
\mathbb{E}_{(p^-, y^-)}\left[\log\frac{\overbrace{\pi_\theta(y^+\mid p^+)}^{\pi_\theta(y^+\mid p^+)\to 1}}{\underbrace{\pi_\theta(y^+\mid p^+)}_{\pi_\theta(y^+\mid p^+)\to 1} + \pi_\theta(y^-\mid p^-)}\right] \\
&= \mathbb{E}_{(p^-, y^-)}[-\log(\pi_\theta(y^-\mid p^-)+1)]
\end{aligned}
\end{equation}
where, $\pi_\theta(y^-\mid p^-)$ is the probability of generating a negative sample $y^-$ for the reference model in stage $i$. 
Throughout the training, the Negative DPO is embedded into SFT stages, forming a DPO-SFT cycle.
The model's staged negative issues are addressed sequentially at each stage via Negative DPO, and then improve the positive capability via SFT (flow matching).

\begin{figure*}[t!]
    \centering
\includegraphics[width=.95\linewidth]{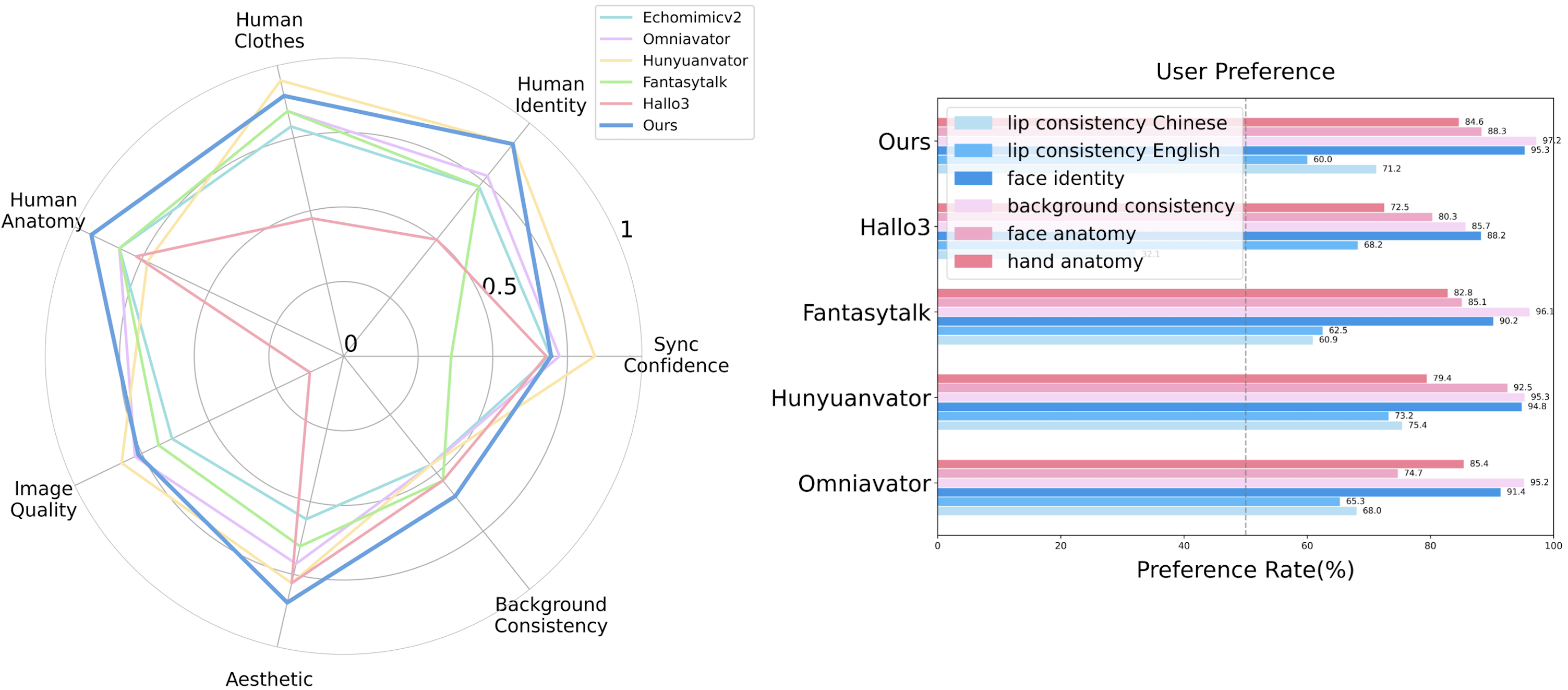}
    \vspace{-0.13in}
    \caption{Evaluation and user study results for SOTA methods.}
    \label{fig:rb}
\end{figure*}

\begin{table*}[t!]
\centering
\resizebox{\textwidth}{!}{
\begin{tabular}{c|cccccccccc|cccc}
\hline
Methods  & Sync-C$\uparrow$ & Sync-D$\downarrow$ & FID$\downarrow$ & FVD$\downarrow$ & IQA$\uparrow$ & ASE$\uparrow$ &ID$\uparrow$&HA$\uparrow$&HC$\uparrow$&BC$\uparrow$ &Head Steps &Human Steps&Head Decay & Human Decay   \\ \hline\hline
EchoMimicV3-1.3B &5.49	&9.67&	42.45&	496.76	&4.91&	3.77	&1.0	&0.95	&0.99	&0.97&5&1min&25&4min \\ \hline
EchoMimicV2-0.8B &5.48	&9.65	&43.72	&543.82	&4.81	&3.34	&0.96	&0.9	&0.97	&0.95&8&4min&8&4min \\ 
Hallo3 &5.42	&9.65	&68.6	&865.32	&4.4	&3.67	&0.91	&0.87	&0.91	&0.96&50&16min&50&16min  \\
FantasyTalk-14B& 4.05	&11.01	&45.03	&603.95	&4.85	&3.48	&0.96	&0.90	&0.97	&0.96&25&18min&25&18min \\
HunyuanAvatar-14B &6.12	&9.11	&42.54	&676.28	&4.96	&3.67	&1.0	&0.85	&1.0	&0.95&50&17min&50&17min \\
OmniAvatar-1.3B &  5.61	&9.58	&53.24	&705.21	&4.92	&3.57	&0.97	&0.90	&0.98	&0.95&50&9min&50&9min  \\ 
\hline
\end{tabular}
}
\vspace{-0.07in}
\caption{Quantitative results for EchoMimicV3.}
\label{tab:1}
\vspace{-0.15in}
\end{table*}

\subsection{Inference Strategy}
\noindent\textbf{Phase-aware Negative classifier-free Guidance (PNG)} is motivated by the observation that the model can effectively reject negative samples after Negative DPO. PNG strengthens the negative prompt outputs of CFG in different timestep phases. Specifically, motion-related negative prompts and detail-related negative prompts are weighted in the early and later phases, respectively, to mitigate undesired artifacts associated with different timestep phases.

\noindent\textbf{Long video inference} predominantly employs a frame-wise sliding window with overlapping frames to extend generated video length in recent work. However, these methods often suffer from unnatural transitions, color discrepancies, and identity inconsistencies across windows. We identify that the devil lies in the details: the computation of the CFG within the overlapping frames requires careful smoothing. Specifically, for each frame in the overlap, our improved Long Video CFG calculation is formulated as follows:
\begin{equation}
\begin{aligned}
\hat{\epsilon}_{\theta}^{w_o}(f) = \epsilon_{\theta}^{w}(f) + s \cdot \left( \textstyle\sum_{i \in \{w, w+1\}} \alpha^i \epsilon_\theta^{i}(f) - \epsilon_{\theta}^{w}(\varnothing) \right)
\end{aligned}
\end{equation}
where $\hat{\epsilon}_{\theta}^{w_o}(f)$ denotes the noise prediction of the $f$ frame of the overlap latents for CFG, $\epsilon_{\theta}^{w}(f)$ and $\epsilon_{\theta}^{w}(\varnothing)$ denote the noise predictions with and w/o conditions for the $f$ frame in the overlap between $w$ and $w+1$ latents window. We compute the weighted average of frames $f$ corresponding to consecutive sliding windows $w$ and $w+1$ as follows:
\begin{equation}
\begin{aligned}
\textstyle\sum_{i \in \{w, w+1\}} \alpha^i \epsilon_\theta^{i}(f) = \left(1-\frac{f}{N}\right) \cdot \epsilon_\theta^{w}(f) + \frac{f}{N} \cdot \epsilon_\theta^{w+1}(f)
\end{aligned}
\end{equation}

\section{Experiments}
\begin{figure*}[t]
    \centering
    \includegraphics[width=0.9\linewidth]{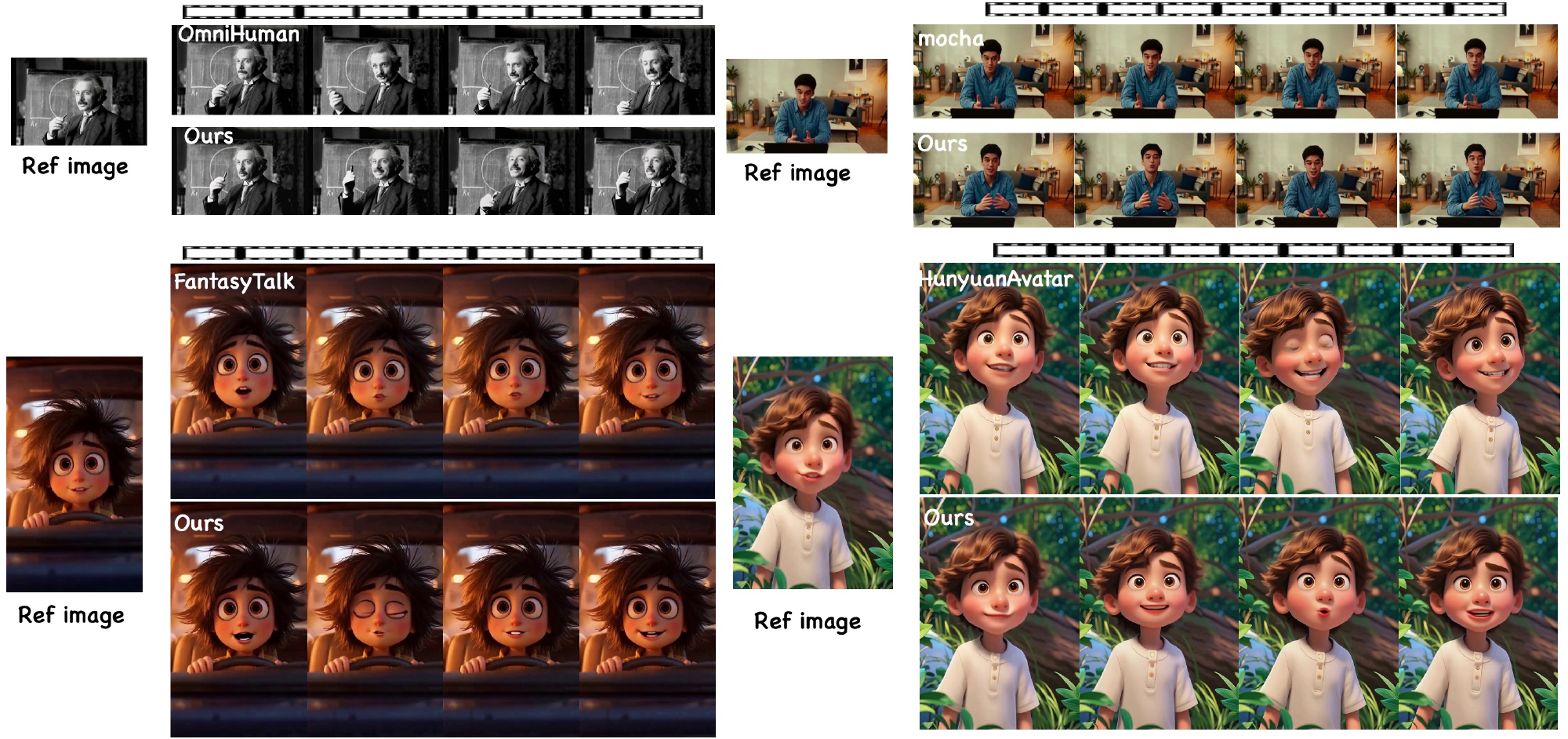}
    \vspace{-0.03in}
    \caption{Qualitative comparison with SOTA methods for talking human animation.}
    \vspace{-0.08in}
    \label{fig:qual}
\end{figure*}
\subsection{Experimental Setups}
\noindent\textbf{Implementation.}
We utilize Wan2.1-FUN-inp-480p-1.3B~\footnote{https://github.com/aigc-apps/VideoX-Fun} as the foundational model. The input video length is set to 113. The classifier-free guidance for text is set to 3, and classifier-free guidance for audio is set to 9. The training process is conducted on 64 96GB GPUs, with a learning rate set to 1e-4. For efficiency, we first extract the VAE latents and caption embeddings for all the training data.

\noindent\textbf{Dataset.}
We train our model on a combination of the EchoMimicV2 dataset, the HDTF dataset~\cite{zhang2021flow}, and additional self-collected data. To ensure data quality, we apply preprocessing steps, including audio synchronization and subtitle removal. The total training data comprises approximately 1,500 hours of video content.

\noindent\textbf{Metrics.} 
To comprehensively evaluate our model's performance, we employ the following metrics:
1) Image Quality: Fréchet Inception Distance (FID)~\cite{deng2019accurate} is used to assess the fidelity and visual quality of generated images.
2) Video Quality: Fréchet Video Distance (FVD)~\cite{unterthiner2018towards} measures temporal coherence and overall video quality.
3) Perceptual and Aesthetic Analysis: We analyze perceptual quality (IQA) and aesthetic appeal (ASE) of the generated content.
4) Audio-Visual Alignment: For lip-sync tasks, we use Sync-C and Sync-D metrics~\cite{prajwal2020lip} to evaluate synchronization accuracy.
5) Consistency Metrics: We adopt Vbench2.0~\cite{zheng2025vbench} metrics, including Identity Consistency (ID), Human Anatomy (HA), Clothing Consistency (HC), and Background Consistency (BC).
For quantitative evaluation, we randomly select 300 videos generated by EchoMimicV3.

\subsection{Results}
\noindent\textbf{Comparison with SOTA Methods.}
We conducted both qualitative and quantitative comparisons with existing methods for half-body digital human generation. These include EchoMimicV2 \cite{meng2025echomimicv2}, HunyuanAvatar \cite{chen2025hunyuanvideo}, OmniAvatar \cite{gan2025omniavatar}, Hallo3~\cite{cui2024hallo3}, and MultiTalk~\cite{sung2024multitalk}. 
As shown in Table~\ref{tab:1}, our method is competitive in various evaluation metrics, such as audio-lip synchronization, human motion accuracy, identity preservation, video aesthetics, and overall video quality, even compared to methods with $10\times$ parameters (e.g., FantasyTalk). Notably, we achieve superior performance in identity preservation, video aesthetics, background consistency, clothing fidelity, and body motion precision over competing methods. 

The normalized results for each dimension are illustrated in Fig.~\ref{fig:rb}, and the corresponding user study validation is also shown within the same figure. Notably, our approach achieves superior Chinese lip-sync accuracy and human motion fidelity compared to state-of-the-art methods. Furthermore, a qualitative comparison with SOTA methods is provided in Fig.~\ref{fig:qual} to demonstrate the effectiveness of our method.

\begin{table}[t!]
\centering
\resizebox{.5\textwidth}{!}{
\begin{tabular}{c|cccccc}
\hline
Methods & Sync-C$\uparrow$ & FVD$\downarrow$ & IQA$\uparrow$ & ASE$\uparrow$ & ID$\uparrow$ & HA$\uparrow$ \\
\hline\hline
EchoMimicV3                  &5.49  &496.76  &4.91  &3.77  &1.00  &0.95  \\
\hline\hline
Task Schedule A              &5.40  &498.09  &4.94  &3.81  &0.97  &0.87  \\
Task Schedule B              &4.98  &499.78  &4.90 &3.73 &0.96 &0.95 \\
w/o EMA                      &5.32  &508.82  & 4.95 &3.87  &0.99  &0.90  \\
\hline\hline
Modals Allocation A          &4.76  &496.90  &4.92  & 3.78 & 1.00 & 0.93 \\
Modals Allocation B          &5.51  &489.08  &4.95  &3.76  &0.97 &0.81  \\
Modals Allocation C          &5.51  &540.80  &4.60  &3.45  &0.91  & 0.93 \\
\hline\hline
SFT only                     &4.94  &540.30  &4.66  &3.65  &0.99  &0.87  \\
SFT+DPO                      &5.21  &480.98  &4.99  &3.82  &0.93  &0.89  \\
\hline\hline
w/o PNG                      &5.49 & 496.07 &4.91  & 3.78 &1.00  &0.89  \\
\hline\hline
w/o Long Video CFG           &5.49  &530.21  &4.77  &3.69  &0.98  &0.92 \\
\hline
\end{tabular}
}
\caption{Ablation study for EchoMimicV3.}
\vspace{-0.12in}
\label{tab:2}
\end{table}

\begin{figure}[t]
    \centering
    \includegraphics[width=\linewidth]{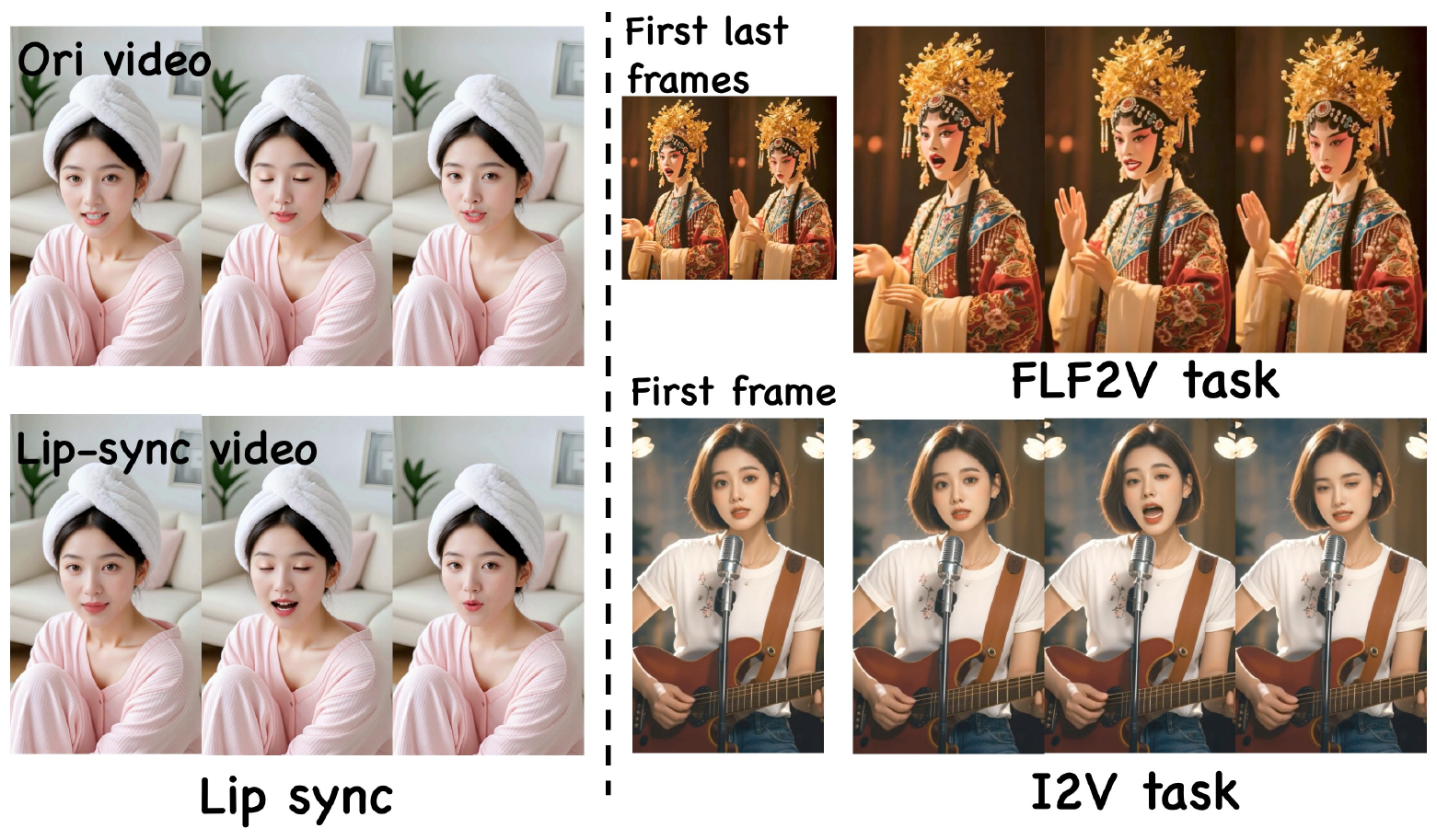}
    \caption{Results for multiple tasks.}
    \vspace{-0.13in}
    \label{fig:multitask}
\end{figure}

\noindent\textbf{Results on Multiple Tasks.}
We validate the multi-task capability of our method. As shown in Fig. \ref{fig:multitask}, it effectively handles diverse scenarios, including lip synchronization (LC), Image-Audio-to-Video (IA2V), and First-and-Last-Frame-Audio-to-Video (FLFA2V). These results demonstrate that our approach is a robust and task-versatile framework, seamlessly integrating multiple task-specific experts into a unified model—a capability that even state-of-the-art methods with up to $10\times$ more parameters fail to achieve.

\subsection{Ablation Studies}
\noindent\textbf{Ablation on Soup-of-Tasks Training Strategy.}
We experimentally validate the efficacy of our training strategy within the Soup-of-Tasks framework, focusing on the counter-intuitive task scheduling and implicit task mixture mechanisms. Specifically, we compare different scheduling approaches (see Table \ref{tab:2}):
Task Schedule A follows an easy-to-hard sequence, starting with lip-sync and progressing to Image-to-Video, while Task Schedule B employs uniform random sampling. Results in the rows $1\sim3$ of Table \ref{tab:2} demonstrate that the counter-intuitive scheduling achieves superior lip synchronization, body motion accuracy, and identity preservation across tasks, whereas alternative schedules yield suboptimal results due to inadequate adaptation of the pretrained model, highlighting the importance of our task allocation strategy in balancing multi-task performance.

We also conduct ablation studies on the inter-task training schedule. It can be observed from row $4$ in Table~\ref{tab:2} that omitting EMA (Exponential Moving Average) negatively impacts performance across tasks, particularly when tasks are jointly trained, leading to motion issues.

\noindent\textbf{Ablation on Multi-Modal PhDA.}
We evaluate different modals allocation strategies across timestep phases. Specifically, Modal Allocation A omits audio in the early phase, Modal Allocation B excludes text from the early/mid phases, and Modal Allocation C omits image from the early phases. Results in Table~\ref{tab:2} show that audio, text, and image modals exhibit distinct phase-specific responses. Deviations from the modality-phase correspondence lead to suboptimal performance: the late phase audio allocation causes lip-sync failures, exclusion of text from the early phases results in motion collapse, and the absence of image in mid-phases impairs identity preservation.

\begin{figure}[t]
    \centering
    \includegraphics[width=\linewidth]{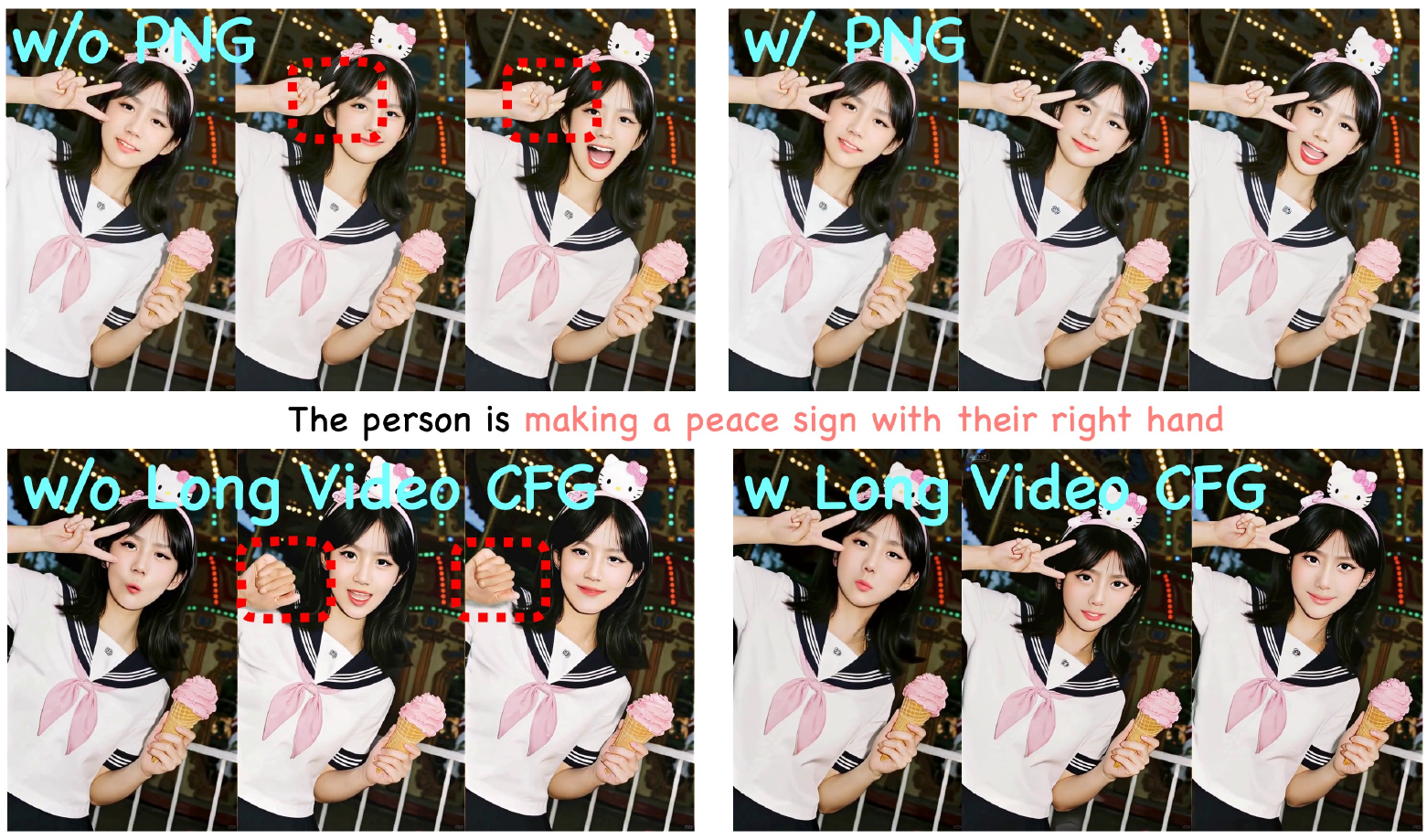}
    \caption{Ablation on PNG and Long Video CFG.}
    \vspace{-0.1in}
    \label{fig:png}
\end{figure}
\noindent\textbf{Ablation on Negative DPO.} 
We study the impact of Negative DPO on the EchoMimicV3 performance. We train our model with only SFT and the traditional paired DPO with the same data as NDPO-SFT cycle.Unfortunately, both SFT-only training and SFT-DPO training generate inferior results compared to the NDPO-SFT cycle training. We observe that SFT-DPO typically suffers from identity consistency issues, as the paired preference data are exclusively sampled from the original training dataset. To incorporate additional paired preference data, substantial human and computational resources would be required; this also demonstrates the paired data-efficiency of 
Negative DPO.

\noindent\textbf{Ablation on PNG.} 
We investigate the impact of the PNG module on human animation performance. As illustrated in Fig.~\ref{fig:png}, the left image depicts the results without the initial 3 timesteps of PNG, where the character's gestures exhibit noticeable degradation compared to the right image that incorporates PNG. These results demonstrate that PNG effectively rejects corresponding negative issues.

\noindent\textbf{Ablation on Long Video CFG.} Long-term video inference without Long Video CFG exhibits overexposure issues in intermediate frames and introduces color discrepancies between consecutive frames. In contrast, incorporating Long Video CFG enables extended-length video generation with natural transitions and prompts following over long durations. The results are illustrated in Fig. \ref{fig:png}. 

\noindent\textbf{Ablation on Efficiency.} Our method achieves talking head generation within 5 inference steps. It can be seen in Table~\ref{tab:1}, compared to 14B-parameter models such as FantasyTalk and HunyuanAvatar, EchoMimicV3 provides a $18\times$ speedup. For talking human synthesis, leveraging TeaCache~\cite{liu2025timestep}, our approach generates a 5-second video in approximately 4 minutes with 25 inference steps.

\section{Related Work}
\subsection{Video Generation}
Early research in video generation focused on 2D U-Net architectures and motion modules, later extended to 3D frameworks. The incorporation of spatiotemporal convolutions enabled models like SVD to produce short clips with high spatial fidelity. Diffusion-based models, such as DiT, further advanced the field by leveraging self-attention mechanisms, reducing inductive bias, and improving temporal coherence and motion fluidity. Recent progress in text-to-video synthesis has been propelled by frameworks like CogVideo~\cite{yang2024cogvideox}, which combine large-scale vision-language pretraining with hierarchical token fusion for precise semantic alignment. HunyuanVideo~\cite{kong2024hunyuanvideo} introduced a dual-stream DiT architecture, enhancing conditional control in complex scenarios. Wan~\cite{wan2025wan} demonstrated scalable models with state-of-the-art performance, setting new benchmarks and highlighting the potential for broader video synthesis applications.

\subsection{Human Animation}
The objective of audio-driven human animation is to synthesize natural, expressive gestures from speech, ensuring semantic, emotional, and temporal alignment. While prior work has focused on animating talking heads~\cite{zhu2024champ, xu2024vasa}, recent methods have advanced quality and consistency. For instance, EMO~\cite{tian2024emo} uses a Frame Encoding module for temporal coherence, and AniPortrait~\cite{wei2024aniportrait} maps 3D facial structures to 2D poses for coherent sequences. Vlogger~\cite{zhuang2024vlogger} generates high-fidelity videos using a diffusion-based framework. EchoMimic~\cite{chen2024echomimic} supports flexible generation modes, driven by audio, facial poses, or both. CyberHost~\cite{lin2024cyberhost} enables versatile animations with multimodal controls, including hand poses and body trajectories. OmniHuman~\cite{lin2025omnihuman} achieves state-of-the-art performance using progressive dropout strategies. Hunyuanvideo-Avatar~\cite{chen2025hunyuanvideo} and MultiTalk~\cite{sung2024multitalk} support multi-audio-driven multi-character animations, while Omniavatar~\cite{gan2025omniavatar} injects audio efficiently into transformers. However, DiT-based methods still struggle with lip sync and long-duration video generation, highlighting the need for further advancements.  %
\section{Conclusion}
In this work, we propose an effective framework, EchoMimicV3, to
master multi-task and multi-modal human animation in a single 1.3B model. We firstly propose a new Soup-of-Tasks paradigm to unify and allocate multiple tasks. Furthermore, we develop a new training strategy called Negative DPO, which is incorporated into the supervised fine-tuning (SFT) process to dynamically mitigate undesired behaviors. We also propose innovative inference strategies, including the Phase-aware Negative-enhanced CFG and the Long Video CFG to enhance vividness and support long-term video generation, respectively.
Extensive experiments demonstrate that EchoMimicV3 remains competitive even when compared to models ten times its model scale.
{
    \small
    \bibliographystyle{ieeenat_fullname}
    \bibliography{main}
}


\end{document}